\documentclass{llncs}
\usepackage{subfig}
\usepackage{graphicx}
\usepackage{amssymb}
\usepackage{ctable}
\usepackage{multirow}
\usepackage{dsfont}
\usepackage{tabularx}

\begin{document}

\title{GridNet with automatic shape prior registration for automatic MRI cardiac segmentation.\vspace{-0.4cm}}
\titlerunning{Deep learning cardiac segmentation}
\author{Cl\'{e}ment Zotti\inst{1} \and Zhiming Luo\inst{1,4} \and Olivier Humbert\inst{3} \and Alain Lalande\inst{2}  \and Pierre-Marc Jodoin\inst{1}}
\authorrunning{Clement Zotti et al.}
\tocauthor{Cl\'{e}ment Zotti, Zhiming Luo, Olivier Humbert, Alain Lalande, Pierre-Marc Jodoin}
\institute{Computer Science Department, Université de Sherbrooke, Canada \(^1\) \\
Le2i, Universit\'{e} de Bourgogne Franche-Comt\'{e}, Dijon France \(^2\) \\
Department of Nuclear Medicine, Centre Antoine Lacassagne, Nice, France \(^3\) \\
Cognitive Science Department, Xiamen University, China \(^4\) \\
\vspace{-0.6cm}}

\maketitle

\begin{abstract}
In this paper, we propose a fully automatic MRI cardiac segmentation method based on a novel deep convolutional neural network (CNN) designed for the 2017 ACDC MICCAI challenge. The novelty of our network comes with its embedded shape prior and its loss function tailored to the cardiac anatomy.  Our model includes a cardiac center-of-mass regression module which allows for an automatic shape prior registration.  Also, since our method processes raw MR images without any manual preprocessing and/or image cropping, our CNN learns both high-level features (useful to distinguish the heart from other organs with a similar shape) and low-level features (useful to get accurate segmentation results).  Those features are learned with a multi-resolution conv-deconv "grid" architecture which can be seen as an extension of the U-Net.

Experimental results reveal that our method can segment the left and right ventricles as well as the myocardium from a 3D MRI cardiac volume in $0.4$ second with an average Dice coefficient of $0.90$ and an average Hausdorff distance of \(10.4\) mm.
\vspace{-0.3cm}

\keywords{Convolutional neural networks, MRI, Heart, segmentation\vspace{-0.5cm}}
\end{abstract}
\vspace{-0.4cm}
\section{Introduction}
\vspace{-0.2cm}

MRI is the gold standard modality for cardiac assessment~\cite{epstein2007mri,giles2009gold}. In particular, the use of kinetic MR images along the short axis orientation of the heart allows accurate evaluation of the function of the left and right ventricles. For these examinations, one has to delineate the left ventricular endocardium (LV), the left ventricular epicardium (or myocardium - MYO) and the right ventricular endocardium (RV) in order to calculate the volume of the cavities in diastole and systole (and thus the ejection fraction), as well as the myocardial mass \cite{peng2016review}. These parameters are mandatory to detect and quantify different pathologies. 

As of today, the clinical use of cardiovascular MRI is hampered by the amount of data to be processed (often more than 10 short axis slices and more than 20 phases per slice). Since the manual delineation of all 3D images is clinically impracticable, several semi-automatic methods have been proposed, most of which being based  on active contours, dynamic programming, graph cut or some atlas fitting strategies~\cite{petitjean2015survey,auger2014semi,peng2016review,grosgeorge2013graph,petitjean2011survey}.  Unfortunately, these methods are far from real time due to the manual interaction which they require. Also, most of them are ill-suited for segmenting simultaneously the LV, the RV, and the MYO.

So far, a limited number of fully-automatic cardiac segmentation methods have been proposed.  While some use traditional image analysis techniques like the Hough transform \cite{wang2015spatiomyo} or level sets \cite{liu2016distance}, fully-automatic segmentation methods are usually articulated around a machine learning method \cite{petitjean2015survey}, and more recently deep learning (DL) and convolutional neural networks (CNN) \cite{phivutran2016fcnet}.  Among the best CNN segmentation models  are those involving a series of convolutions and pooling layers followed by one \cite{long2015fully} or several \cite{noh2015learning} deconvolution layers.  In 2015, Ronneberger et al. \cite{ronneberger2015u} proposed the U-Net, a CNN which involves connections between the conv and deconv layers and whose performances on medical images are astonishing. Recently, DL methods have been proposed to segment cardiac images \cite{phivutran2016fcnet,tan2016cardiac,ngo2017combining}. While Tran \cite{phivutran2016fcnet} applied the well-known fCNN \cite{long2015fully} on  MR cardiac images, Tan et al. \cite{tan2016cardiac} used CNN to localize (but not segment) the LV, and Ngo et al. \cite{ngo2017combining} use deep belief nets again to localize but not segment the LV. 
These methods are doing only one or two class segmentation and do not incorporate the shape prior inside the network.  

In this paper we propose the first CNN method specifically designed to segment the LV, RV and MYO without a third party segmentation method.  Our approach incorporates a shape prior whose registration on the input image is learned by the model.

\vspace{-0.3cm}
\section{Our Method}
\vspace{-0.2cm}
The goal of our method is to segment the LV, the RV and the MYO of  a 3D $N\times M\times H$ raw MR image $X$. This is done by predicting a 3D label map $T$ also of size $N\times M\times H$ and whose voxels $\vec v= (i,j,k)$ contain a label $T_{\vec v} \in \{\mbox{Back, LV, RV, MYO}\}$, where "Back" stands for tissues different than the other three.  Following the ACDC structure, X is a series of short axis slices starting from the mitral valve down to the apex~\cite{Kastler2004} (please refer to the ACDC website for more details~\cite{ACDC}).  In order to enforce a clinically plausible result, a shape prior $S$ is provided which encapsulates the relative position of the LV, RV and MYO.  The main challenge when using a shape prior such as this one is to align it correctly onto the input data X~\cite{tavakoli2013survey}.  Since the size and the orientation of the heart does not vary much from one patient to another, we register $S$ on $X$ by translating the center of $S$ on the cardiac center of mass (CoM) $\vec{c}$ of $X$.  The CoM is computed based on the location of the pericardium (obtained from MYO and RV) in each slice.  Since $\vec c$ is not provided with the input image, our method has a regression module designed to predict it.  To our knowledge, our approach is the first to incorporate a shape prior as well as its registration within an end-to-end trainable structure.

\vspace{-0.3cm}
\subsection{Shape prior}
\vspace{-0.2cm}
The shape prior $S$ is a 3D volume which encodes the probability of a  3D location $\vec v=(i,j,k)$ of being part of a certain class (Back, LV, RV, or MYO).  We estimate this probability by computing the pixel-wise empirical proportion of each class based on the groundtruth label fields $T_i$ of the training dataset:\vspace{-0.3cm}
\begin{eqnarray}
P({\cal C}|\vec v)=\frac{1}{N_t}\sum_{i=1}^{N_t}  \mathds{1}_{\cal C} \left(T_{i,\vec v} \right) \nonumber
\end{eqnarray}
where $\mathds{1}_{\cal C} \left(T_{i,\vec v} \right)$ is an indicator function which returns 1 when $T_{i,\vec v}=\cal C$ and 0 otherwise, and $N_t$ is the total number of training images.  

These probabilities are put into a $3 \times 20 \times 100 \times 100$ volume $S$ where $3$ stands for the 3 classes (RV, MYO, LV)\footnote{No need to store the probability of "Back" since the 4 probabilities sum up to 1.}, $20$ stands for the number of interpolated slices (from the base to the apex) and $100 \times 100$  is the inplane size.  Note that prior to compute $S$, we realign the CoM of all training label fields $T_i$  into a common space and crop a \( 100 \times 100 \) region around that center. 

\subsection{Loss}

The goal of our system is to predict a correct label field $T$ given an input image $X$ while automatically aligning the shape prior $S$ on $X$ by aligning their center of masses $c$.  In order to do so, our loss incorporates the following four terms:

\begin{eqnarray}
\label{eq:loss}
	\mathcal{L} = \small{\sum_i}  \Big(
    \underbrace{- \gamma_T \sum_{l=1}^{4} \sum_{\vec v} T_{i,l,\vec v}  \ln \hat{T}_{l,\vec v}}_{\mathcal{L}_T}
    \underbrace{- \gamma_C \sum_{l=1}^{4} \sum_{\vec v} C_{i,l,\vec v}  \ln \hat{C}_{i,l,\vec v} }_{\mathcal{L}_C} + 
    \underbrace{\gamma_c ||c_{i,\mathbf{w}}-\hat c_i||^2}_{\mathcal{L}_c} \Big) + 
    \underbrace{\gamma_w ||\vec w||^2}_{\mathcal{L}_w} \nonumber
\end{eqnarray}

Here $\mathcal{L}_T$ and $\mathcal{L}_C$ are the cross-entropies of the predicted labels and the predicted contours. In this equation, $l$ stands for the class index, $\vec v$ is a pixel location, and \(\gamma_T\) and $\gamma_C$ are constants. \( T_{i,l,\vec v} \) is the true probability that pixel $\vec v$ is in class \(l\), and \( \hat{T}_{i,l,\vec v} \) is the output of our model for pixel $\vec v$ and class $l$ while  \(C_i\) and \(\hat{C}_i\) are contours extracted from \(T_i\) and \(\hat{T}_i\).  Note that the use of a contour loss has been shown by Luo et al~\cite{Luo2017} to enforce a better precision.  As for $\mathcal{L}_c$, it is the Euclidean distance between the predicted CoM $c_{i,\mathbf{w}}$ and the true CoM, and $\mathcal{L}_w$ is the prior loss.

\begin{figure}[tp]
\centering
\includegraphics[width=\textwidth,height=180px]{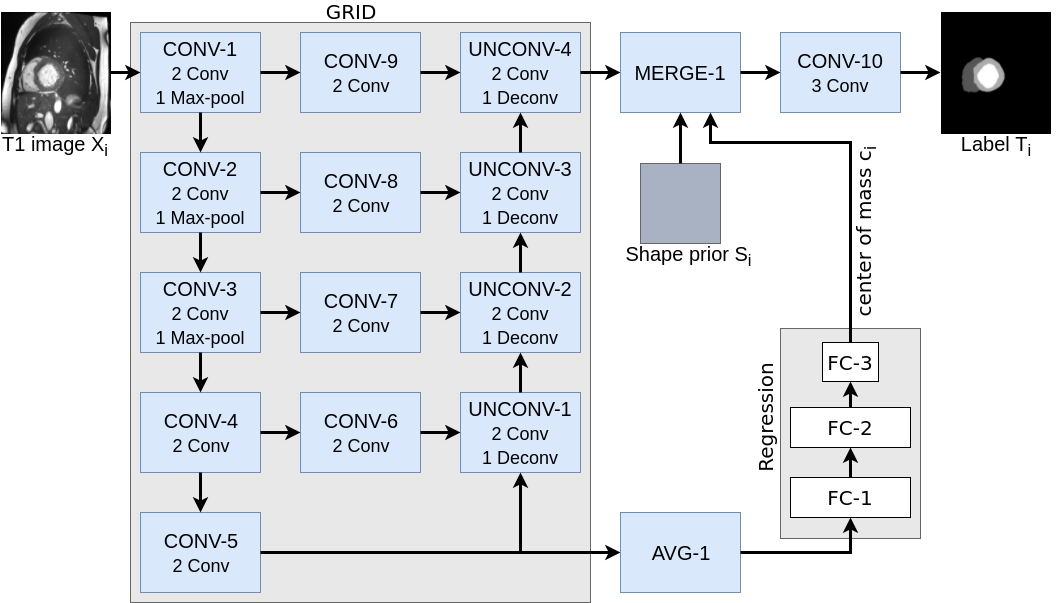}
\caption{Our network architecture.}
\label{fig:ournet}
\vspace{-4mm}
\end{figure}

\vspace{-0.3cm}
\subsection{Proposed network}
\vspace{-0.2cm}
The goal of our CNN is to learn good features for predicting the label field $T_i$ as well as the  CoM $\vec c_i$ which is used to align the shape prior $S$ on the input image $X_i$.  In other words, a network that has the ability of minimizing the loss of Eq.(\ref{eq:loss}).  With that objective, good features must account for both the global and the local context: the global context to differentiate the heart from the surrounding organs and estimate its CoM, and the local context to ensure accurate segmentation and prediction of contours.  

In that perspective, we implemented a grid-like CNN network with 3 columns and 5 rows (c.f. Fig.~\ref{fig:ournet}).  
The input to our model (upper left) is an $ 256 \times 256 $ MR image $X_i$ and the shape prior $S_i$ of the corresponding slice, while the output is the CoM $\vec c_i$ (bottom right) and a label field $T_i$ (top right) also of size $ 256 \times 256$.  Note that a common issue with MRI cardiac images is the fact that along the 2D short-axis, the location of the heart sometimes get shifted from one slice to another due to different breath-holds during successive acquisitions.  As a consequence, instead of processing a 3D volume as a whole, we feed the network with 2D slices as shown in Fig.~\ref{fig:ournet} and reshape the 3D volume $T_i$ by stacking up the resulting 2D label fields.

As we get  deeper in the network (from  CONV-1 to CONV-5), the extracted features involve a larger context of the input image.  Since the CONV-5 layer includes high-level features from the entire image, we use it to predict the cardiac CoM $\vec c_i$  of the input image $X_i$.  The second column contains 4 convolution layers (all without max-pooling)  used to compute features at various resolutions.  The last column aggregates features from the lowest to the highest resolution.  The UNCONV-4 layer contains both global and local features which we use to segment the image.   Note that this grid structure is similar to the U-Net except for the middle CONV-6 to 9 layers and the fact that we use the CONV-5 features to estimate a CoM $\vec c_i$.
Each conv layer has a \(3 \times 3\) receptive field and its feature maps have the same size than their input (zero padding).  We also batch normalize each feature map, use the ReLU activation function, and dropout \cite{srivastava2014dropout} to have a better generalization. Please note that so instead of having \(32\) millions parameters like the U-Net, our gridNet has approximately \(8\) millions parameters.

\vspace{-0.5cm}
\subsubsection{Estimating the center of mass $\vec c_i$}
The CoM $\vec c_i$ is estimated with a regression module located after the CONV-5 layer. An average pooling layer (AVG-1) is used to reduce the number of features fed to the FC-1 layer. The FC-1, FC-2 and FC-3 layers are all fully connected and the output of FC-3 is the (\(x_i, y_i\)) prediction of $\vec c_i$.

\vspace{-0.5cm}
\subsubsection{Estimating the label field $T_i$}
The output of the UNCONV-4 layer has 4 $256\times 256$ feature maps which we append to the shape prior $S$.  The MERGE-1 layer realigns $S$ based on the estimated CoM $\vec c_i$ and use zero padding to make sure $S$ has a $256\times 256$ inplane size.  In this way, the output of MERGE-1 has 7 feature maps: 4 from UNCONV-4 and 3 from $S$.  The last CONV-10 layer is used to squash those 7 feature maps down to a 4D output.

\vspace{-0.5cm}
\subsubsection{Training}  The model is trained by minimizing the loss $\cal L$ of Eq.(\ref{eq:loss}). We use a batch size of 10 2D MR images taken from the ED or ES phase independently and the ADAM optimizer~\cite{kingma2014adam}.  The model is trained with a learning rate of $10^{-4}$ for a total of 100 epochs.  

\vspace{-0.5cm}
\subsubsection{Pre and Post processing}
The input 3D images \(X\) are pre-processed by clamping the 4\% outlying grayscale values, zero-centering the grayscales and normalize it by their standard deviation.  Once training is over, we remove  outliers by keeping the largest connected component for each class on the overall predicted 3D volume.  Note that these pre and post processes are applied to every model tested in Section~\ref{sec::res}.

\vspace{-0.3cm}
\section{Experimental Setup and Results}
\label{sec::res}
\vspace{-0.2cm}
\subsection{Dataset, evaluation criteria, and other methods}
\label{sec::data}
\vspace{-0.1cm}

Our system was trained and tested on the 2017 ACDC dataset.  Since the testing dataset was not available as of the paper submission deadline, we trained our system on 75 exams and validated it on the remaining 25 exams.  The exams are divided into 5 evenly distributed groups: dilated cardiomyopathy, hypertrophic cardiomyopathy, myocardial infarction with altered left ventricular ejection fraction, abnormal right ventricle and patients without cardiac disease. 

Cine MR images were acquired in breath hold with a retrospective or prospective gating and with a SSFP sequence in 2-chambers, 4-chambers and in short axis orientations.  A series of short axis slices cover the LV from the base to the apex, with a thickness of 5 to 8 mm and often an interslice gap of 5 mm. The spatial resolution goes from 0.83 to 1.75 mm$^2$/pixel.  For more details on the dataset, please refer to the ACDC website~\cite{ACDC}.

In order to gauge performances, we report the clinical and geometrical metrics used in the ACDC challenge.  The clinical metrics are the correlation coefficients for the cavity volume and the ejection fraction (EF) of the LV and RV, as well as correlation coefficient of the myocardial mass for the End Diastolic (ED) phase.  As for the geometrical metrics, we report the Dice coefficient~\cite{zou2004statistical} and the Hausdorff distance~\cite{huttenlocher1993hausdorff} for all 3 regions and phases. 

We compared our method with two recent CNN methods:  the conv-deconv CNN by  Noh et al.~\cite{noh2015learning} and the U-Net by  Ronneberger et al.~\cite{ronneberger2015u}.  We chose those methods based on their excellent segmentation capabilities but also because their architecture can be seen as a particular case of our approach.

\begin{table}[tp]
\centering
\caption{Results for the validation dataset.}
\begin{tabularx}{\textwidth}{@{\extracolsep{\fill} } l|cc|cc|cc}
    \specialrule{.1em}{.05em}{.05em} 
    & \multicolumn{2}{c}{Dice LV} & \multicolumn{2}{c}{Dice RV} & \multicolumn{2}{c}{Dice MYO} \\ \hline
    			& ED			& ES			& ED			& ES			& ED			& ES \\ \hline 
    ConvDeconv	& 0.92			& 0.87			& 0.82			& 0.64			& 0.76			& 0.81 \\ \hline
    UNet 		& \textbf{0.96}	& 0.92			& 0.88			& 0.79			& 0.78			& 0.76 \\ \hline
    Our method	& \textbf{0.96}	& \textbf{0.94}	& \textbf{0.94}	& \textbf{0.87}	& \textbf{0.89}	& \textbf{0.90} \\ \hline \hline

& \multicolumn{2}{c}{HD LV (mm)} & \multicolumn{2}{c}{HD RV (mm)} & \multicolumn{2}{c}{HD MYO (mm)} \\ \hline
    	 		& ED			& ES			& ED				& ES				& ED			& ES \\ \hline
    ConvDeconv	& 8.77			& 10.34			& 22.59				& 28.45				& 13.92			& 11.64 \\ \hline
    UNet 		& 6.17			& 8.29			& 20.51				& 21.20				& 15.25			& 17.92 \\ \hline
    Our method	& \textbf{5.96}	& \textbf{6.57}	& \textbf{13.48}	& \textbf{16.66}	& 8.68	& \textbf{8.99} \\ 
    \specialrule{.1em}{.05em}{.05em} 
\end{tabularx}

\begin{tabularx}{\textwidth}{@{\extracolsep{\fill} } l|c|c|c|c|c}
	\specialrule{.1em}{.05em}{.05em} 
     				& Corr EF LV 		& Corr EF RV		& Corr MYO ED		& Corr LV vol	& Corr RV vol \\ \hline
    ConvDeconv		& 0.988				& 0.764				& 0.927				& 0.990			& 0.957 \\ \hline
    UNet 			& 0.991				& 0.824				& 0.921 			& 0.995			& 0.821 \\ \hline
    Our method		& \textbf{0.992}	& 0.898				& \textbf{0.975}	& \textbf{0.997}& \textbf{0.978} \\ 
    \specialrule{.1em}{.05em}{.05em}
\end{tabularx}


\label{tab:geopatho}
\vspace{-0.2cm}
\end{table}

\vspace{-0.3cm}
\subsection{Experimental results}
\vspace{-0.2cm}
As can be seen in Table~\ref{tab:geopatho}, our method outperforms both the conv-deconv and the U-Net. The Dice coefficient is better by an average of \(5 \)\%, and the Hausdorff distance is lower for our method by an average of $4.4$ mm.  This is a strong indication that the contour loss and the shape prior help improving results, especially close to the boundaries.  Without much surprise, the RV is the most challenging organ, mostly because of its complicated shape, the partial volume effect close to the free wall, and intensity inhomogeneities. The clinical metrics are also in favor of our method.   Based on the  correlation coefficient, our method is overall better than conv-deconv and UNet, especially for the myocardium (bottom of Table~\ref{tab:geopatho}).

\begin{figure}[tp]
\centering

\includegraphics[width=\textwidth]{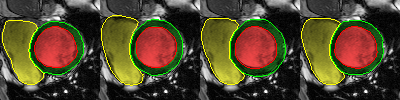}\vspace{-0.4mm}
\includegraphics[width=\textwidth]{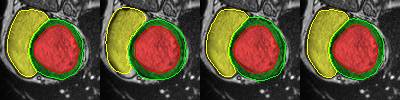}\vspace{-0.4mm}

\includegraphics[width=\textwidth]{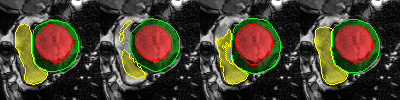}\vspace{-0.4mm}

\includegraphics[width=\textwidth]{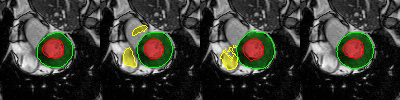}\vspace{-0.4mm}
\includegraphics[width=\textwidth]{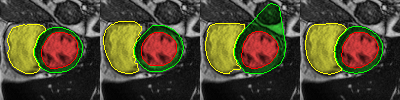}\vspace{-0.4mm}

\includegraphics[width=\textwidth]{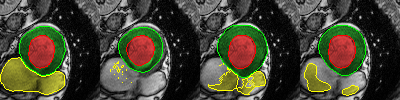}\vspace{-0.4mm}
\caption{From left to right: ground truth, conv-deconv, U-Net, and our method.\vspace{-0.3cm}}
\label{fig:res}
\end{figure}

Careful inspection reveal that errors are not uniformly distributed.  Interestingly, conv-deconv and U-Net produce accurate results on most slices of each 3D volume as illustrated in the first two rows of Fig.~\ref{fig:res}.  That said, they often get to generate a distorted result for 1 or 2 slices (out of 7 to 17) which end up decreasing the Dice score and increasing the Hausdorff distance. 
This situation is shown in rows 3,4, and 5 of Fig.~\ref{fig:res}.  Overall, the right ventricle is the most challenging region for all three methods.   It is especially true at the base of the heart, next to the mitral valve where the RV is connected to the pulmonary artery.  This is illustrated in the last row of Fig.~\ref{fig:res}.

\vspace{-0.4cm}
\section{Conclusion}
\label{sec::conclu}
We proposed a new CNN method specifically designed for MRI cardiac segmentation.  It implements a "grid" architecture which is a generalization of the U-net. It  uses a shape prior which is automatically aligned on the input image with a regression method.  The shape prior forces the method to produce anatomically plausible results. Experimental results reveal that our method produces an average DICE score of 0.9.  This shows that an approach such as our's can be seen as a decisive step towards a fully-automatic clinical tool used to compute functional parameters of the heart. In the future, we plan on generalizing our method to other modalities (such as echocardiography or CT-scan) as well as other organs such as the brain.  In that case, we shall propose a more elaborate registration module which shall implement an affine transformation.\vspace{-0.5cm}
%
%
{\footnotesize

\bibliographystyle{ieeetr}
\bibliography{miccaibib}

}
\end{document}